\RequirePackage{fix-cm}
\documentclass[smallextended]{svjour3}       
\smartqed  
\usepackage{graphicx}
\begin{document}

\title{Swarm Programming Using Moth-Flame Optimization and Whale Optimization Algorithms
}

\titlerunning{Swarm Programming Using MFO and WOA}        

\author{Tapas Si         
}


\institute{T. Si \at
              Department of Computer Science \& Engineering\\
              Bankura Unnayani Institute of Engineering\\
              Bankura-722146, West Bengal, India\\
              \email{c2.tapas@gmail.com.com}           
}

\date{Received: date / Accepted: date}

\maketitle

\begin{abstract}
Automatic programming (AP) is an important area of Machine Learning (ML) where computer programs are generated automatically. Swarm Programming (SP), a newly emerging research area in AP, automatically generates the computer programs using Swarm Intelligence (SI) algorithms. This paper presents two grammar-based SP methods named as Grammatical Moth-Flame Optimizer (GMFO) and Grammatical Whale Optimizer (GWO). The Moth-Flame Optimizer and Whale Optimization algorithm are used as search engines or learning algorithms in GMFO and GWO respectively. The proposed methods are tested on Santa Fe Ant Trail, quartic symbolic regression, and 3-input multiplexer problems. The results are compared with Grammatical Bee Colony (GBC) and Grammatical Fireworks algorithm (GFWA). The experimental results demonstrate that the proposed SP methods can be used in automatic computer program generation.
\end{abstract}

\section{Introduction}
\label{sec:intro}
Automatic programming~\cite{Rich} is a machine learning technique by which computer programs are generated automatically in any arbitrary language. SP~\cite{Olmo} is an automatic programming technique which uses SI algorithms as search engine or learning algorithms. The grammar-based SP is a type of SP in which \textit{Context-free Grammar} (CFG) is used to generate computer programs in a target language. Genetic Programming (GP)~\cite{Koza} is a evolutionary algorithm in which tree-structured genome is used to represent a computer program and Genetic algorithm (GA) is used as a learning algorithm. Grammatical Evolution (GE)~\cite{Ryan},\cite{Neill1} is a variant of grammar-based GP (GGP)~\cite{Mckay} in which linear genome i.e., array of integer codons is used to represent a genotype and Backus-Naur Form (BNF) of CFG is used to generate the computer programs (i.e., phenotype) from the genotype. Generally, variable-length GA is used as a learning algorithm in GE. SP uses GP-like tree-structure genome which represents a computer program and SI algorithms are used as learning algorithms.
O.~Roux and C.~Fonlupt~\cite{Roux} proposed ant programming in which ant colony optimizer (ACO)~\cite{Dorigo} was used to generate the computer programs. 
D.~Karaboga et al.~\cite{Karaboga} proposed artificial bee colony programming (ABCP) for symbolic regression and artificial bee colony (ABC) algorithm~\cite{Karaboga2} was used as learning algorithm.
A.~Mahanipour and H.~Nezamabadi-pour~\cite{Mahanipour} proposed Gravitation Search Programming (GSP) in which Gravitation Search Algorithm (GSA)~\cite{Rashedi} was used as a learning algorithm.
The grammar-based SP are the variants of GE where SI algorithms are used as search engines or learning algorithms through genotype-to-phenotype mapping using BNF of CFG. Grammatical Swarm (GS)~\cite{Neill2},\cite{Neill3},\cite{Neill4}, Grammatical Bee Colony (GBC)~\cite{Si1}, and Grammatical Fireworks algorithm (GFWA)~\cite{Si2} are grammar-based SP. Particle Swarm Optimizer (PSO)~\cite{Kennedy} was used as search engine in GS. ABC algorithm was used as search engine in GBC and Fireworks algorithm (FWA)~\cite{Tan} was used as search engine in GFWA. Till date, as per author's best knowledge, Moth-Flame Optimization (MFO)~\cite{MFO} and Whale Optimization algorithm (WOA)~\cite{WOA} are not used in automatic programming. Therefore, in the current work, two grammar-based SP algorithms GMFO and GWO are proposed. In GMFO algorithm, MFO is used as a search engine to generate computer programs automatically through genotype-to-phenotype mapping using BNF of CFG. Similar to GMFO, WOA is used as a search engine in GWO for automatic computer program generation. The proposed two methods are applied to Santa Fe Ant Trail, symbolic regression and 3-input multiplexer problem. The results of GMFO and GWO are compared to the results of GFWA and GBC. The experimental results demonstrates that the proposed two methods can be used in automatic computer program generation in any arbitrary language.

\section{Materials \& Methods}
\label{sec:1}
In this work, two grammar-based SP such as GMFO and GWO are proposed. In GMFO, MFO algorithm is used as a search engine or learning algorithm to generate computer programs automatically through genotype-to-phenotype mapping. In GMFO, each individual is an array of integer codons in the range $[0,255]$ and it represents the genotype and derived computer program from genotype using BNF of CFG is known as phenotype. First of all, it is primary concern to define problem specific CFG in BNF. An example of CFG in BNF is given below:

\begin{verbatim}
1. <expr> := (<expr><op><expr>) (0) | <var> (1)
2. <op> :=  + (0) | - (1) | * (2) | / (3) 
3. <var> := x1 (0) | x2 (1)
 \end{verbatim}
The $i$th individual of the search engine, i.e., set of $d$ integer codons $X_i (x_1, x_2, \ldots, x_d)$ are initialized as follows:
\begin{equation}
 x_j=round(255 \times rand (0,1))
 \end{equation} 
A \textit{mapping process} maps the rule numbers from the codons in the derivation process of computer programs in the following way:
rule=(codon integer value) MOD (number of rules for the current non-terminal)

\begin{figure}[!bth]
  \includegraphics[scale=0.6]{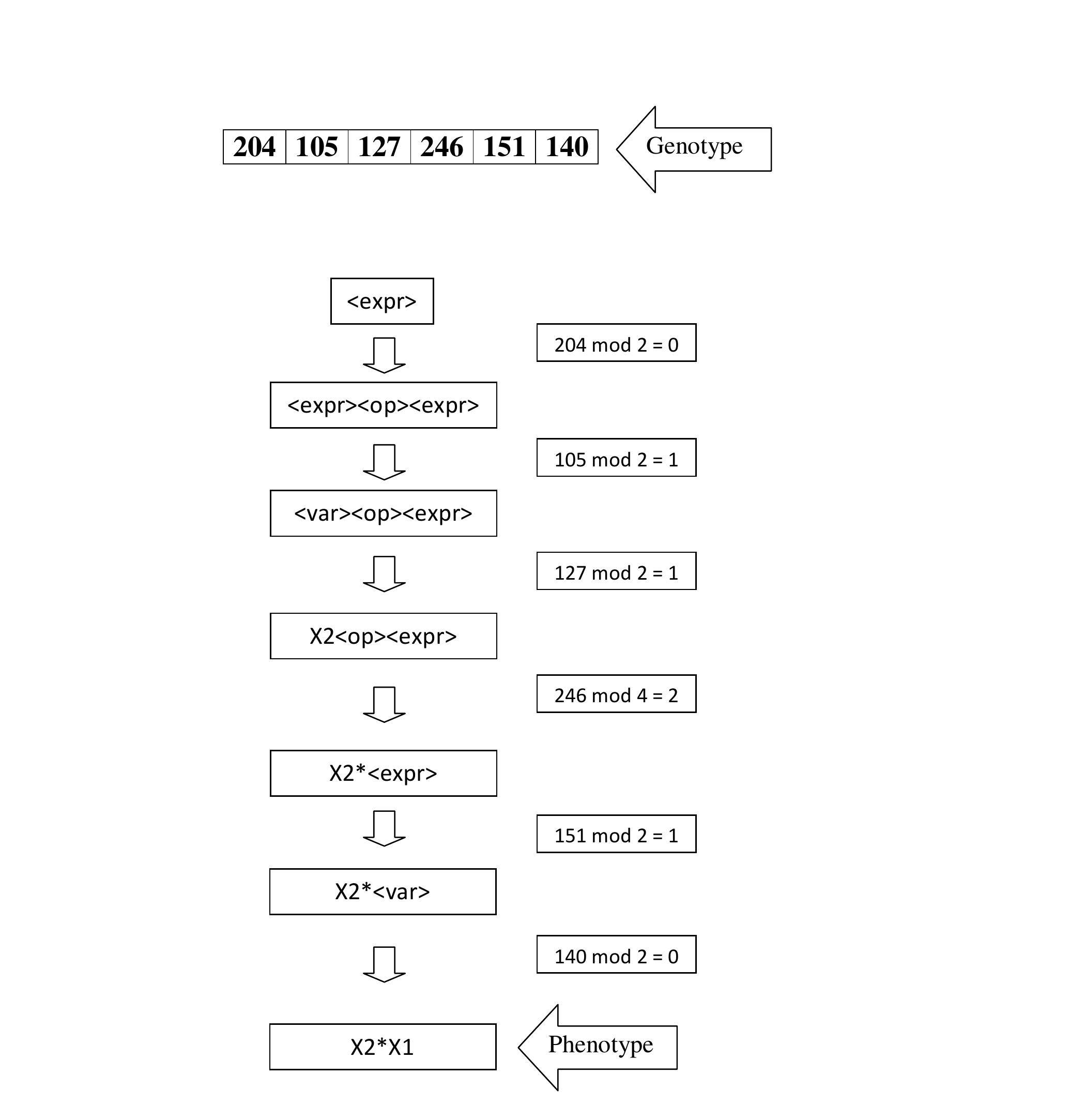}
\caption{Genotype-to-phenotype mapping process}
\label{fig:1}       
\end{figure}

The representation of genotype, phenotype and genotype-to-phenotype mapping are given in Figure~\ref{fig:1}.
Similar to GFMO, the positions of whale are the set of integer codons in the range $[0,255]$. The same genotype-to-phenotype mapping process as in GMFO is used to generate the computer programs. During the derivation process, if the derivation is run out of codons, then the process restarts from the beginning. This process is called as \textit{wrapping}. After a certain number of wrapping, if there is any non-terminal remain in the derived program, then the corresponding individual is denoted as invalid. The invalid individual is replaced by the valid individual later on during the search process.

Search engine or learning algorithms are another important part of GMFO and GWO. As already mentioned before, MFO and WOA are used as search engines in GMFO and GWO respectively. MFO algorithm is based on the transverse orientation, i.e., the navigation method of moths in nature. Moths fly in the night by maintaining a fixed angle with respect to the moon for travelling in a straight line for long distance. But they are trapped in a useless or deadly  spiral around the artificial light source. This phenomena is modelled in MFO algorithm. The detail description of MFO algorithm can be obtained from~\cite{MFO}. WOA is another nature-inspired meta-heuristic algorithm which mimics the social behaviour of humpback whales. The humpback whales live alone or in group and their favourite  prey is krill and small fish herds. Their foraging behaviour, i.e., bubble-net feeding method is done by creating bubbles along a circle. The Bubble-net attacking method (i.e., exploitation phase) has two steps namely shrinking encircling mechanism and spiral updating position. The searching mechanism for prey is used to create exploration of the search space. The detail of WOA can be obtained from~\cite{WOA}.

%
 \section{Experimental Setup}
 \label{sec:4}
\subsection{Benchmark Problems}
Three benchmark problems such as Santa Fe Ant Trail (SFAT), symbolic regression, and 3-input multiplexer are chosen for the experiment. The objective of the SFAT problem is to find out the program by which an ant can eat 89 piece of food placed in $32 \times 32$  grid in 600 time steps. The target function for symbolic regression problem is $f(x)=x+x^2+x^3+x^4$ and $100$ fitness cases are generated randomly in the range [-1,1]. $3$-input multiplexer problem has  $8$ fitness cases. The detail descriptions and problem specific defined grammar can be obtained from ~\cite{Si1},\cite{Si2}.

\subsection{Parameters Settings}
\label{sec:param}

The parameters  of GMFO are set as the following: number of search agents ($\mathcal{N}$) = $30$,
 dimension = $100$.
The parameters  of GWO are set as the following: number of search agents ($\mathcal{N}$) = $30$,
 dimension = $100$. The other parameters in GMFO and GWO are dynamically controlled as in~\cite{MFO},\cite{WOA}
As the functions are not evaluated for invalid individual in the above algorithms, the maximum number generations  or iterations are not fixed for the comparative study. Therefore,
each of GFW, GBC, GMFO, and GWO algorithms is allowed to run for maximum $30,000$ number of function evaluations (FEs) in a single run. All algorithms are terminated when they reach the maximum FEs or target error. The target errors are set to 0, 0.01 and 0 for ant trail, symbolic regression and 3-multiplexer problems respectively. The numbers of wrapping are set to $3$, $2$ and $1$ for ant trail, symbolic regression and 3-multiplexer problems respectively.

\subsection{PC Configuration}
\begin{itemize}
\item Operating System: Windows 10 Pro
\item CPU: Intel(R) Core(TM) i7-9700K @3.6GHz
\item RAM: 64 GB
\item Software: Matlab 2018a
\end{itemize}

\section{Results \& Discussion}
\label{sec:results}
The proposed GMFO and GWO algorithms are applied to Santa Fe Ant Trail (SFAT), symbolic regression, and 3-input multiplexer problems. The experiments are repeated for 30 times independently for each algorithms. The mean and standard deviation of best-run-errors over $30$ independent runs are given in Table~\ref{tab:1}. The number of successful runs and success rates (in \%) over $30$ independent runs are given in Table~\ref{tab:2}. The success rate (SR) is calculated as follows:
\[SR=\frac{\textit{number of achieving the target}}{\textit{number of total runs}}\]
The number of successful runs and success rates (in \%) over $30$ independent runs are given in Table~\ref{tab:mean_fes}.
The results of GFWA and GBC are obtained from study~\cite{Si2} as the current work is the part of the same project.

\begin{table}[!bth]
  \caption{Mean and standard deviation of best-run-errors over $30$ independent runs.}
 \label{tab:1}
 
 \begin{tabular}{   lrrr  }
 \hline\noalign{\smallskip}
    Algorithms &  Santa Fe Ant Trail &  Symbolic Regression &  3-Multiplexer\\
\noalign{\smallskip}\hline\noalign{\smallskip}
GFWA & $ 24.57 (16.9516)$ & $\textbf{6.65 (7.246)}$ & $0.93 (0.2537)$\\ 
GBC & $32.57 (17.8609)$ & $10.35 (7.1404)$ & $\textbf{0.7(0.4661)}$\\ 
GMFO& $35.53 (22.8212)$ & $10.35 (7.814)$ & $0.8 (0.407)$\\
GWO & $\textbf{23.57(17.2880)}$ & $10.15(7.4338)$ & $1.00 (0.00)$\\ 
\noalign{\smallskip}\hline
\end{tabular}

\end{table}

\begin{table}[!bth]
  \caption{Number of successful runs and success rates (in \%) over $30$ independent runs. }
 \label{tab:2}
 \begin{tabular}{lrrr}
 \hline\noalign{\smallskip}
    Algorithms &  Santa Fe Ant Trail &  Symbolic Regression &  3-Multiplexer\\
\noalign{\smallskip}\hline\noalign{\smallskip}
GFWA & $1 (3.33\%)$ & $\textbf{15(50.00\%)}$ & $2 (6.67\%)$\\ 
GBC & $0 (0.00\%)$ & $7 (23.33\%)$ & $\textbf{9(30.00\%)}$\\ 
GMFO & $\textbf{2(6.67\%)}$ & $9 (30.00\%)$ & $6 (20.00\%)$\\ 
GWO & $1 (3.33\%)$ & $9 (30.00\%)$ & $0 (0.00\%)$\\
\noalign{\smallskip}\hline
\end{tabular}
\end{table}

\begin{table}[!h]
 \caption{Mean and standard deviation of FEs over $30$ independent runs }
 \label{tab:mean_fes}
  \begin{tabular}{lrrr}
 \hline\noalign{\smallskip}
    Algorithms & Santa Fe Ant Trail &  Symbolic Regression &  3-Multiplexer\\
\noalign{\smallskip}\hline\noalign{\smallskip}
GFWA & $ 29917(453.88)$ & $ 23943(8657.80)$ & $29062(4415.30)$\\ 
GBC & $30000(0.00)$ & $27076(6935.20)$ & $\textbf{23549(10420.00)}$\\ 
GMFO & $\textbf{28224.9(6812.3395)}$ & $\textbf{21586.9(13097.6074)}$ &$26200 (9020)$ \\
GWO &  $29553.8  (2524.6001)$ & $22432.07 (12023.5602)$ & $30000(0.00)$\\
\noalign{\smallskip}\hline
\end{tabular}
\end{table}
From Table~\ref{tab:1}, it is observed that the GWO performs better than other algorithms for SFAT problem. GFWA performs better than other algorithms for symbolic regression problem. GBC performs better than others for 3-multiplexer problem. If the results of GMFO are compared with GWO, it can be observed that GWO provides a higher accuracy than GMFO for SFAT and symbolic regression problems whereas GMFO provides higher accuracy than GWO only for 3-multiplexer problem.

From Table~\ref{tab:2}, it is observed that the success rate of GMFO is higher than all others algorithms for SFAT problem. GFWA provides higher success rate than others for regression problem whereas GBC provides higher success rate than others for multiplexer problem. If the success rates of GMFO and GWO are compared, then it can be observed that GMFO has higher success rate than GWO for SFAT and  multiplexer problems and there is a tie for regression problem. 

From Table~\ref{tab:mean_fes}, it is observed that the mean FEs taken by GMFO is lower than other algorithms for SFAT and regression problems whereas the mean FEs taken by GBC is lower than others for multiplexer problem.
The computer programs evolved by GMFO and GWO are given below: \\ 
The successful ant program evolved by GMFO (ant eats all $89$ pieces of food): 
\begin{verbatim}
  if(foodahead()) if(foodahead()) if(foodahead()) 
  if(foodahead())   if(foodahead()) move(); else 
  if(foodahead()) if(foodahead())  left();  else
  if(foodahead()) if(foodahead()) left(); else 
  left(); end; else   if(foodahead()) right(); 
  else if(foodahead())  if(foodahead()) if(foodahead())
  left(); else left(); end;   else move(); end; else 
  move(); end; end; end; end;   else   if(foodahead())
  if(foodahead()) if(foodahead()) if(foodahead()) 
  if(foodahead()) right(); else left(); end; else left();
  end; else   move(); end;  else left(); end; else move();
  end; end; end; else   right(); end; else   if(foodahead())
  move(); else move(); end; end; else left(); end; else
  left();  end; move(); left(); if(foodahead()) move(); 
  else if(foodahead()) right(); else right(); end; end; right();
\end{verbatim}
 \medskip
\noindent

The ant program evolved by GWO (ant eats $88$ out of $89$ pieces of food): 
\begin{verbatim}
 if(foodahead()) left(); else right(); end; right(); 
 if(foodahead()) move(); else left(); end; move(); left();
\end{verbatim}
 \medskip
\noindent
A successful program evolved by GMFO for symbolic regression problem (absolute error = $1.7837e-15$):
\begin{verbatim}
plus(times(x,plus(times(x,plus(x,times(x,times(pdivide(x,x),x)))),
x)),x)
 \end{verbatim}
    \medskip
\noindent
A successful program evolved by GWO for symbolic regression problem (absolute error = $4.6668e-15$):
\begin{verbatim}
times(plus(plus(times(minus(times(x,x),pdivide(x,x)),x),x),x),
plus(x,pdivide(x,x)))
 \end{verbatim}
    \medskip
\noindent

A successful program evolved by GMFO for 3-multiplexer problem (absolute error = 0):
\begin{verbatim}
nor(nor(nor(x3,x1),nor(nor(x1,nor(x1,x1)),x2)),nor(nand(nor(x2,x1),
nor(x1,x1)),x3))
 \end{verbatim} 
\medskip
\noindent

A program evolved by GWO for 3-multiplexer problem (absolute error = 1):
\begin{verbatim}
nand(or(x3,x1),x2)
 \end{verbatim} 

From the discussion of the results, it is found that no single algorithm performs better than all other algorithms for all problems in this study. The above presented results are the experimental evidence of the fact that the proposed GMFO and GWO algorithms can be used in automatic computer program generation in any arbitrary language. 

\section{Conclusion}
This paper presents two grammar-based swarm programming methods namely GMFO and GWO. The proposed methods are applied to solve SFAT, symbolic regression, and 3-input multiplexer problems. The experimental results demonstrate that the proposed methods can be used to generate computer programs automatically in any arbitrary language. In this  study, the basic version of MFO and WOA are utilized as search engines or learning algorithms in genotype-to-phenotype mapping. The update version of these algorithms can be used to obtain better performance in automatic computer program generation. In the future, the GMFO and GWO can be applied to real-world problems such as data classification and regression analysis.

%
%


\begin{thebibliography}{}
%
%

\bibitem{Rich}
Rich C, Waters R.C (1998) Automatic Programming: Myths and Prospects, IEEE Computer,
21(8): 40–51

\bibitem{Koza}
 Koza J.R (1992) Genetic Programming: On the Programming of Computers by Means of
Natural Selection, MIT Press.

\bibitem{Olmo}
Olmo J.L, Romero J.R, Ventura S.(2014) Swarm-based metaheuristics in automatic programming:
a survey, WIREs Data Mining Knowl Discov 2014. doi: 10.1002/widm.1138

\bibitem{Ryan}
 Ryan C, Collins J.J, O’Neill M (1998) Grammatical Evolution: Evolving Programs for an
Arbitrary Language, In: BanzhafW, Poli R, Schoenauer M, Fogarty T.C. (eds.) EuroGP 1998.
LNCS, vol. 1391, Springer, Heidelberg, 1998 :83–95

\bibitem{Neill1}
O’Neill M, Ryan C (2001) Grammatical Evolution, IEEE Transactions on Evolutionary Computation
5(4):349–358

\bibitem{Mckay}
6. Mckay R.I, Hoai N.X, Whigham P.A, Shan Y, O’Neill M (2010) Grammar-based genetic
programming: a survey, Genetic Programming and Evolvable Machines, 11: 365–396.

\bibitem{Roux}
Roux O, Fonlupt C (2000) Ant programming: or how to use ants for automatic programming.
In: International Conference on Swarm Intelligence (ANTS); 121–129.

\bibitem{Dorigo}
Dorigo M, Maniezzo V, Colorni A (1996) Ant system: optimization by a colony of cooperating
agents. IEEE Trans Syst Man Cybern B Cybern, 26:29–41. doi:10.1109/3477.484436.

\bibitem{Karaboga}
Karaboga D, Ozturk C, Karaboga N, Gorkemli B (2012) Artificial bee colony programming
for symbolic regression. Inform Sci, 209:1–15. doi: 10.1016/j.ins.2012.05.002.

\bibitem{Karaboga2}
Karaboga D (2005) An Idea Based On Honey Bee Swarm For Numerical Optimization, In:
Technical Report-TR06, Erciyes University, Engineeing Faculty, Computer Engineering Department

\bibitem{Mahanipour}
Mahanipour, A, Nezamabadi-pour, H (2019) GSP: an automatic programming technique with
gravitational search algorithm. Appl Intell 49, 1502–1516 doi:10.1007/s10489-018-1327-7

\bibitem{Rashedi}
Rashedi E, Nezamabadi-Pour H, Saryazdi S (2009) GSA: a gravitational search algorithm.
Inf Sci 179:2232–2248



\bibitem{Neill2}
O’Neill M, Brabazon A (2004) Grammatical swarm, In: Genetic and Evolutionary Computation
Conference (GECCO) :163–174.

\bibitem{Neill3}
 O’Neill M, Brabazon A (2006) Grammatical Swarm: The Generation of Programs by Social
Programming, Natural Computing 5(4): 443–462

\bibitem{Neill4}
O’Neill M, Leahy F, Brabazon A (2006) Grammatical swarm: a variable-length particle
swarm algorithm, Swarm Intelligent Systems, Studies in Computational Intelligence.
Springer: 59–74.

\bibitem{Kennedy}
 Kennedy J, Eberhart R (1995) Particle Swarm Optimization, In: IEEE International Conference
on Neural Networks, Perth, Australia







\bibitem{Si1}
Si T, De A, Bhattacharjee A.K (2013) Grammatical Bee Colony, In: B.K. Panigrahi et al.
(Eds.): SEMCCO 2013, Part I, LNCS 8297:436–445.

\bibitem{Si2}
Si T (2016) Grammatical Evolution Using Fireworks Algorithm, In: M. Pant et al. (eds.),
Proceedings of Fifth International Conference on Soft Computing for Problem Solving, Advances
in Intelligent Systems and Computing 436, DOI 10.1007/978-981-10-0448-3 4.

\bibitem{Tan}
Tan Y, Zhu Y (2010) Firework Algorithm for Optimization, In: Y. Tan et al.(Eds): ICSI 2010,
Part I, LNCS 6145: 355–364, Springer-Verlag Berlin Heidelberg.

\bibitem{MFO}
Mirjalili S (2015) Moth-Flame Optimization Algorithm: A Novel Nature-inspired Heuristic Paradigm, Knowledge-Based Systems (2015), doi: http://dx.doi.org/10.1016/j.knosys.2015.07.006

\bibitem{WOA}
Mirjalili S,  Lewis A (2016) The Whale Optimization Algorithm, Advances in Engineering Software, 95, pp.~51--67.

\end{thebibliography}


\end{document}